# Renewal Strings for Cleaning Astronomical Databases

**Amos J. Storkey***
School of Informatics
5 Forrest Hill
University of Edinburgh

**Nigel C. Hambly**
Institute for Astronomy
Blackford Hill
University of Edinburgh

**Christopher K. I. Williams**
School of Informatics
5 Forrest Hill
University of Edinburgh

**Robert G. Mann**[†]
Institute for Astronomy
Blackford Hill
University of Edinburgh

## Abstract

Large astronomical databases obtained from sky surveys such as the SuperCOSMOS Sky Surveys (SSS) invariably suffer from spurious records coming from artefactual effects of the telescope, satellites and junk objects in orbit around earth and physical defects on the photographic plate or CCD. Though relatively small in number these spurious records present a significant problem in many situations where they can become a large proportion of the records potentially of interest to a given astronomer. We have developed renewal strings, a probabilistic technique combining the Hough transform, renewal processes and hidden Markov models which has proven highly effective in this context. The methods are applied to the SSS data to develop a dataset of spurious object detections, along with confidence measures, which can allow this unwanted data to be removed from consideration. These methods are general and can be adapted to any other astronomical survey data.

## 1 INTRODUCTION

Digital sky surveys in astronomy are a fundamental research resource (e.g. [9]). Surveys form the basis of statistical studies of stars and galaxies, enabling work ranging in scale from the solar neighbourhood to a significant fraction of the observable universe. Surveys are carried out in all wavelength ranges, from high energy gamma rays to the longest wavelength radio atlases. Despite this diversity, there are certain features common to most digital surveys: pixel images at a given spatial and spectral resolution are pro-cessed using a pixel analysis engine to generate lists of object detections containing parameters describing each detection. In most cases, the object detection algorithm has to be capable of finding a heterogeneous family of objects, for example point-like sources (stars, quasars); resolved sources (e.g. galaxies) and diffuse, low surface-brightness, extended objects (e.g. nebulae). Object parameters describing each detection typically include positions, intensities and shapes. The volume of pixel data required to be processed necessitates totally automated pixel processing, and of course no imaging system is perfect.

---
*also at Institute for Astronomy, Edinburgh
[†]also at National E-Science Centre, Edinburgh

This paper looks at a class of problems which are the most significant sources of unwanted records in the SuperCOSMOS Sky Survey (SSS) data. The SSS is described in a series of papers ([4] and references therein). Briefly, the SSS consists of Schmidt photographic plates scanned using the fast, high precision microdensitometer SuperCOSMOS [5]. The survey is made from 894 overlapping fields in each of three colours (blue, red and near-infrared denoted by the labels J, R and I respectively), one colour (R) is available at two epochs to provide additional temporal information. Each image contains approximately $10^9$ 2-byte pixels. The pixel data from each photograph in each colour and in each field are processed into a file of object detections; each object record contains parameters describing that object. The number of object detections on a plate varies from about half a million to ten million records. Presently, the entire southern hemisphere is covered, primarily using plates from the UK Schmidt Telescope. Data and many more details are available online at http://www-wfau.roe.ac.uk/sss.

How spurious features translate into objects in the sky survey catalogue depends also on the approach of the program which processes the digital picture into object catalogue records. For the SuperCOSMOS Sky Surveys many of the largest linear features are clearly non-astronomical in origin, cannot be processed by the pixel analyser, and therefore do not give rise to spurious object catalogue records. The rest tend to be represented in the catalogue as a number of objects lying along a line. Hence even if a track traversed the whole plate in the original image, in the derived catalogue data it might only translate into a set of objects traversing a short section of the original track.

The focus of this paper is on locating objects in an astronomical dataset derived from or affected by satellite tracks, aeroplane tracks or scratches, and how can we accurately distinguish them from true astronomical objects. Because much work has usually already been done deriving the object data from images, because in many cases original image data may not be available, and because of the huge size of the images involved, we are not considering working with the images directly, only with the derived datasets.

## 1.1 RELATED WORK

In [3] and [12] the authors use the Hough transform to locate satellite tracks in astronomical data. Briefly, the Hough transform searches through a finite number of angles, mapping each data point to one of a finite number of lines at that angle. The number of points mapped to each line is counted, and lines with a higher than expected count are flagged as possible tracks. However our initial tests showed the Hough transform was not able to reliably detect the short linear features or partial tracks that are commonly seen in sky survey data. The analysis of Section 4.1 compares the approach of this paper with a Hough transform approach, and verifies these initial tests.

There is a fair body of vision literature on robust techniques for line segmentation. For example in [8] the authors use a smoothing of the Hough accumulator [7] to obtain a robust fit. However these approaches tend to be global straight line methods, and hence they do not work well for either short line segments or curved lines. Cheng, Meer and Tyler [2] provide methods for dealing with multiple structures which need not cover the whole space. Their approach is confined to situations where there is not dominant background data, or large numbers of outliers. Image based techniques for line extraction are common, but tend to be based on continuity considerations, and they are not appropriate in the context where we might be working with data derived from images rather than the images themselves. The important work of Hastie and Stuetzle [6] on principal curves provides a different direction which models curved data, but once again does not provide the robustness and efficiency needed for situations when curves are set in large amounts of other data. Many other more complicated vision approaches to this problem are ruled out because of the sheer amount of data which needs to be considered.

## 2 SPURIOUS OBJECTS

A number of distinct classes of spurious object commonly occur in optical and near-infrared sky survey data. The descriptions given here refer to the form they take within the SSS data. However most other astronomical databases have similar characteristics.

### 2.1 SATELLITE AND AEROPLANE TRACKS

Satellite tracks are due to movement of the satellite over the duration of exposure for a given field. Movement into or out of the Earth's shadow, the two ends of exposure, or removal by the object recogniser can all stop the data related to a satellite track from traversing the whole plate. The positions of satellite tracks are unpredictable, and using a (probably incomplete) catalogue of satellites and orbiting debris would be a complicated and probably unreliable way of locating them. For some narrow and faint tracks the data can be sparsely distributed along the track. For bolder tracks the data might consist of objects with ellipses aligned along the track.

Aeroplane tracks come from aeroplane lights as they cross the field of view. Often (but not always) the lights are flashing, resulting in dashed tracks. All the representational issues which apply to satellite tracks also apply to the data derived from aeroplane lights. Figure 1 gives a few examples of different tracks on the SuperCOSMOS Sky Survey plates and the data (shown as ellipses) which is derived from those plates.

### 2.2 SCRATCHES

Scratches on the plate surface are not uncommon despite all the effort taken to protect the emulsion from such. These scratches can be seen by the SuperCOSMOS digitiser as darker regions and hence are confused with photographic exposure. They are usually (but not always) short, they tend to be curved, and sometimes the curvature can vary significantly along the scratch. Again the same issues occur in translating these linear features into data. An example of a scratch can be seen in Figure 1.

### 2.3 HALOS AND DIFFRACTION SPIKES

Because survey observations are optimised for faint objects, bright stars and galaxies often have optical artefacts associated with them. There can be bright halos from internal reflections within the telescope, along with horizontal and vertical diffraction spikes emanating from the bright star, due to diffraction about the telescope struts. Similar methods to those outlined in

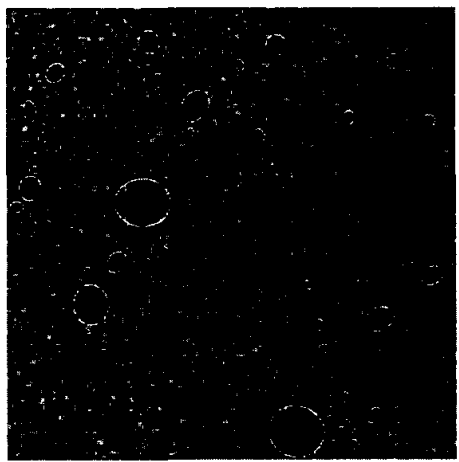
(a)

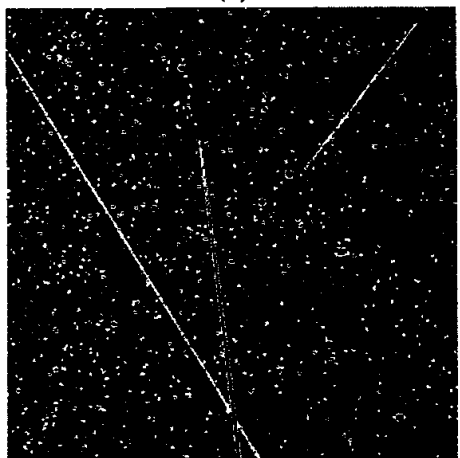
(b)

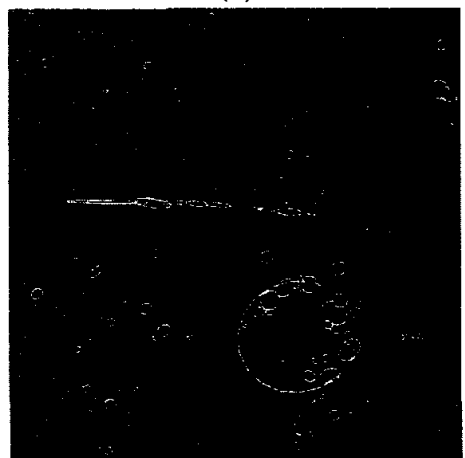
(c)

Figure 1: (a): A faint satellite track with sparse spurious objects distributed along it. (b) A number of aeroplane tracks in field UKJ413. (c) A small slightly curved scratch. Scratches can be longer, fainter or more curved than this, but small scratches are common.

this paper have been tailored for detecting these artefacts. In this paper the focus will be on detection of the other linear features, and the results for halos and diffraction spikes will appear in illustrations but will not be discussed.

## 2.4 PROBLEM OF SPURIOUS OBJECTS

Spurious objects introduce errors in statistical results derived from the data, and make locating particular classes of objects harder. Many tracks result in spurious, elliptical objects with low surface brightness that resemble galaxies, contaminating the respective object catalogue. A single-colour galaxy catalogue, created from the UKJ survey for the purposes of studying faint blue galaxies would therefore be highly contaminated by spurious, aligned image records. This could severely impact a statistical analysis of the type described in [1], where the degree and scale of real galaxy alignment is being sought. In many general problems we may be interested in real objects which might be in one dataset but not in an other, such as objects which are evident at one wavelength but not another. Fast moving stars will also be in different places in catalogues derived from observations at different times, meaning that they will not have exact positional matches across the datasets, e.g. [10]. Unfortunately satellite track artefacts have the same characteristics; they will only ever appear (in the same place) in one dataset, and not in any other. Searches on non-matching objects will bring up all the objects of interest *plus* all of these artefacts. When searching for rare objects the spurious records can be overwhelming. Removing spurious objects, then, is of broad importance in astronomy.

## 3 RENEWAL STRINGS

Renewal strings are a probabilistic data mining tool for finding subsets of records following unknown line segments in data space which are hidden within large amounts of other data. The method was developed specifically to address the problem of this paper. Renewal strings combine a model for two dimensional background data and a set of models for small numbers of data points lying on one dimensional manifolds within the two dimensional space. The design of the model allows efficient line based techniques to be used for separating the background data from the different one dimensional manifolds. In this section we define a generative renewal string model and then show how the key variables can be inferred from the *real* data. Although this inference is approximate, it captures the fundamental characteristics of the model.

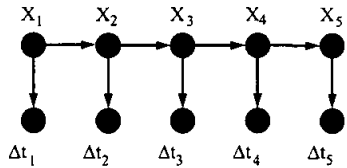

Figure 2: Belief network for the renewal process HMM.

### 3.1 RENEWAL PROCESS HMMs

The techniques developed in this paper utilise hidden Markov models (HMMs) and renewal processes. A renewal process is a model for event times obtained by defining a probability distribution for the time between events (commonly termed the inter-arrival time). The time at which event $i$ occurs is dependent only on the time of the previous event $i-1$, that is it has the Markov property. In this paper we will be looking at the distance between points on a line rather than the time between events, and hence this distance will be called the inter-point distance.

It is possible to assume the inter-point distance is the visible output from a hidden Markov model. Figure 2 provides the belief network for this combination. Here the $\Delta t$'s label the inter point distances along the line, and the $X$'s denote the class label of the hidden Markov model. Different class labels will represent different renewal process models for the observables. In this setting the different classes will represent different classes of tracks and a 'background' class corresponding to the normal density of stars and galaxies.

The combination of renewal processes and hidden Markov models is not new within temporal settings. For example, in the case that the renewal processes are all Poisson processes, there is a direct relationship between the Renewal Process hidden Markov model and the Markov modulated Poisson process [11].

### 3.2 RENEWAL STRINGS

The Renewal String generative model is built as follows. First 2 dimensional star and galaxy positions are generated from a background spatial model. This could be any spatial process such as an inhomogeneous Poisson process. For the purposes of this paper we define the background model to be a Poisson process which is homogeneous within small regions, but has different rates in different regions. Denote this rate function $\Lambda(\mathbf{r})$ for positions $\mathbf{r}$.

Track processes are superimposed on the background data, to simulate satellite tracks or scratches. There are potentially a number of different track classes, each with different inter-point distributions. The tracks are generated as follows. For each $\theta$ from a large but finite set of angles $\Theta$, and for each of a finite set of lines $L$ at that angle, each of a given (narrow) width $w$, a renewal process HMM is used to generate track data. Along each line, a Poisson process (with large mean inter-point distance $\gamma$) is used as a *birth* process for the track; an event in this process signifies the start of the track generation. The class of the track is chosen from the prior distribution $P(X_0)$, and track points are sampled by generating from a renewal process HMM: the inter-point distance is sampled from $P(\Delta t_i | X_i)$ conditioned on the current class, and a new point $t_{i+1}$ is placed the distance $\Delta t_i$ away from the current point, at the given angle from that point. Then the next class is chosen from the transitions $P(X_{t+1} | X_t)$.

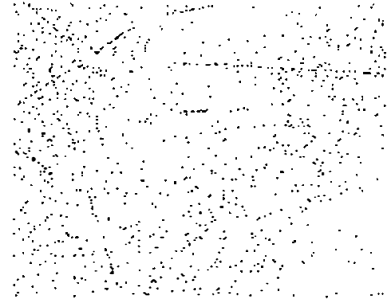

Figure 3: A sample from a 3 hidden-state renewal strings prior, illustrating a background model, and 7 lines with differing characteristics.

We stop generating the track either when the edge of the plate is reached or the hidden Markov chain transitions into the 'stop' class. The transition into the stop class initiates the birth process again, which allows more than one track to be generated along the same line. Note it is possible for a track to be turned on before reaching the region of interest (in this case the plate edge), but not yet be turned off, and hence the track will start at the edge of the plate. As the birth process produces rare events most lines will not contain any tracks at all. Finally each point in each track is independently perturbed perpendicular to the line of the track uniformly across the track width $w$.

Figure 3 illustrates a sample from a generative model of this form. We see a background model, along with two different types of tracks, one of which is a high density broken line, the other a medium density line. The model only generates straight line segments. Curves can be approximated using piecewise linear segments.

### 3.3 INFERENCE AND LEARNING

We require a reasonably fast inference scheme for this model. Borrowing from the Hough transform approach it could be sensible to resort to line based techniques in order to perform inference. The Hough transform looks through a comprehensive set of lines in the data, and finds those with a high accumulator. To implement the renewal string we take this one step further. Rather than just count the points along the line, a renewal process HMM is run along the line to find points which could be best classified as part of a track. This approach is only an approximate inference scheme for the aforementioned generative model. The main issue is that, as with the Hough transform, the dependence between lines at different angles is ignored. The inference scheme for a single line is exact in the case where data from all tracks other than those along that line have been removed. In reality, though, such data remains. However because there are few tracks, tracks at other angles will contribute at most a small number of points to the data along the current line, and so this is likely to have limited effect on the inference for the current line. This is the primary approximation assumption of the inference method.

To work with lines rather than with spatial variables, we use the fact that a spatial point distribution which is an inhomogeneous Poisson process will correspond to an inhomogeneous Poisson process along the length of any line (with some given width) going through that region of space. Hence when we condition on the fact that we are considering one particular line, a one dimensional Poisson process can be used instead a spatial one. The inhomogeneity of the Poisson process takes care of the fact that the background model is not likely to take the same form across the whole plate.

Suppose we have an estimate for the density $D_b$ of background objects local to each point. The full initialisation and inferential process can now be given:

1 Set the width $w$ based on the expected maximum width of the lines to be found. Define the interpoint distance distribution $P(\Delta t|X, D_b)$ for each class $X$ including the background class. This can depend on the background object density at that point. Define the class transition probabilities $P(X_i^k|X_{i-1}^k)$ and initial probabilities $P(X_0^k)$.

2 For each angle $\theta$ from the set $\Theta$

a For every point in the dataset, find all the lines $L'$ of width $w$ in $L$ which contain the point. Store the position $t$ along each line in $l'$ in a bin corresponding to that line.

b For each line in $L$, sort all the distances in its bin. Use these distances as the data for an HMM with emission probabilities $P(\Delta t|X, D_b)$ and transition probabilities $P(X_i^k|X_{i-1}^k)$. Run the usual forward-backward inference to get marginal posterior class probabilities for each point. Flag any points which have a low probability of being background objects and note the angle at which these points were detected.

At the end of this process, the flagged points are the points suspected of being part of a track or scratch. The associated probability gives extra information regarding the certainty of this classification.

Note that, in terms of the generative model, the transition probability out of the background state $P(X_i^k \neq 0|X_{i-1}^k = 0)$ is given by the probability that the point is generated by the birth process rather than the background process. In practice, at least for this work, we approximate this by a fixed empirically determined value. Then we can take the initial class probability $P(X_0^k)$ as given by the equilibrium distribution of the Markov chain. The set of angles $\Theta$ is generally chosen to be regularly spaced between 0 and 180 degrees, and the lines $L$ are chosen to cover the region of consideration with a 2 line overlap; each point in the space lies in 2 and only 2 lines at a given angle.

To estimate the rate $\Lambda(r)$ of the background inhomogeneous Poisson process, we assume there is a length scale $s$ such that, for regions of size $s \times s$, the contributions from the satellite tracks to the total number of points, and the variation in background star/galaxy density are both negligible. Then the local mean of the background Possion process can be approximately obtained from the total density of points in a local region of size $s \times s$.

Tuning of the parameters could be done with the usual $M$ step of the Baum-Welsh/expectation maximisation algorithm for HMMs. On the other hand empirical ground truth estimates could be used to set the parameters. In this work the tracks are also modelled as Poisson processes (a specific form of renewal process). The fundamental reason for this is that along the line of a satellite track there will also be objects corresponding to stars and galaxies. The point density along a track from a satellite moving in front of a dense distribution of stars will be higher than one passing in front of a relatively sparse region of sky, and hence the line of objects along each track is a superposition. Poisson processes have the advantage that the superposition of two Poisson processes is also a Poisson process. The equivalent statement is not true for more general forms of renewal processes.

## 4 SPURIOUS DATA DETECTION

The renewal strings model was tested on plate datasets within the SSS. For the background star/galaxy pro-cess the local density was obtained by gridding the whole space into 40,000 boxes and counting the elements in each box. 1000 different angle settings were used, and 18000 different lines for each angle. The width $w$ of the lines was chosen based on the largest expected track widths in the data, and the number of lines was chosen to provide coverage at this width. The angular variation was then chosen such that any significant length of any track will not be missed between two different angles.

A simple model of two hidden states was used, one corresponding to the background, another to the satellite track. The inter-point histogram for the satellite track was set to be an exponential distribution using the empirical mean from a training set including 30 different satellite tracks from low density plates (the resulting mean was 360 microns on the plate, corresponding to 24 arcsec on the sky). As stars and galaxies also appear along satellite tracks, this empirical mean was added to the mean of the background process to properly model the density along a satellite track in different circumstances. The transition probabilities were set approximately using prior knowledge about the number of satellite tracks etc. on the training plates, the number of objects in total and the number of objects per satellite track. This resulted in the transition matrix $P(X_t|X_{t-1})$ for $X = \{background, track\}$ of

$$\begin{pmatrix} 0.999998 & 0.04 \\ 2\text{e-}06 & 0.96 \end{pmatrix}$$

Ellipse alignment along a track can provide useful information, and so an alignment state was provided as an input to the HMM, whose parameters were also set empirically, based on the number of aligned rather than unaligned objects which occurred along a track, compared with what would occur along a general line.

### 4.1 HOUGH TRANSFORM COMPARISON

The Hough transform does a different job from the renewal string, as it is designed to find lines which traverse the whole plate. If we wish to find line segments we have to do some post processing of the results. The exact position of the tracks would still need separating from the other points in that Hough box. Even so we can assess how well the Hough transform can find lines which contain the linear features.

There are many decision functions which can be used with the Hough transform. For a useful comparison with the renewal string results, we look at the significance level which would be needed to detect each track that was detected with the renewal string method, assuming a null hypothesis of a Poisson process background distribution, and that lines corresponding to tracks will have larger accumulators than that suggested by the null hypothesis. We also look at how many other false positive tracks would also be detected for given significance levels. The number of angles and line widths considered were set to match the renewal string settings (1000 angles, 9000 different perpendicular distances).

Results from this analysis for plate UKR002 are shown in Table 1. This plate has no satellite tracks that traverse the whole plate, but does have some tracks which are a quarter of plate width. Each track/scratch was located in a semi-automated way and, for analysis purposes, diffraction spikes were ignored by removing all tracks within 1.5 degrees of the horizontal or vertical. The position and angle of each track was noted, and included in a track list. In general each track was noted once, however where there was a large curvature to a track, more than one reference could have been included in the list. Detections relating to a halo about a bright star were removed by hand. This left 35 tracks or scratches in the reference list. All these tracks were detected by the renewal string.

Table 1 shows the significance level required to detect the tracks along with the total number of lines that would be flagged as tracks by the Hough transform at each level. Once again these counts exclude Hough accumulators corresponding to lines within 1.5 degrees of the horizontal or vertical, meaning a total of 968 different angles were considered. Accumulators with an expected count below 12 were ignored; they are easily affected by isolated points.

Many of the tracks are picked up by the Hough transform for high significance levels. However some tracks are not even detectable at significance levels of 0.5 and smaller. Hence the renewal string approach is certainly increasing the detection rate compared with using the Hough transform alone. Furthermore the Hough transform produces large numbers of false positives even when only choosing very significant lines; the number of false positives on this plate is much greater than the theoretical number that should be found. Some of these will be contributions from accumulators mapping to lines overlapping a track at a slight angle. However a dominant reason for the discrepancy is that approaches like the Hough transform do not easily deal with variations in the background density; there is an assumption of homogeneity. If many stars are clustered in one location they can cause a significant contribution to a Hough accumulator.

## 5 EVALUATION

The detections were evaluated by an astronomer looking through a printed version of the plate data for a whole plate (UKR001). The plate was spit into 36 regions, each region being printed out. These printouts were examined closely for false negative and false positive detections. As the measured characteristics of true stars or galaxies along or very near a satellite track will be affected by the track, these objects should also be flagged. Any objects which are labelled as having a probability of being a track greater than 0.5 are flagged. False positive and negative rates will be affected by this choice. Astronomers requiring more stringent criterion could use a higher threshold.

| SIGLEV | DET | TOT | THEOR |
|---|---|---|---|
| 0.5 | 31 | $3.18 \times 10^6$ | $4.4 \times 10^6$ |
| 0.9 | 21 | $8.96 \times 10^5$ | $8.7 \times 10^5$ |
| 0.99 | 9 | $1.62 \times 10^5$ | $8.7 \times 10^4$ |
| $1 - 10^{-4}$ | 5 | 5903 | 871.2 |
| $1 - 10^{-6}$ | 3 | 257 | 8.7 |
| $1 - 10^{-7}$ | 2 | 71 | 0.87 |

Table 1: The number of the 35 tracks/scratches on UKR002 which would have been detected using the Hough transform. SIGLEV gives the significance level used. DET gives the number of the tracks which would have been flagged at that significance level, TOT the total number of lines flagged as significant by the Hough transform, and THEOR the theoretical number of false positives for a homogeneous Poisson distribution. A significance level of $1 - 10^{-7}$ is needed to reduce the theoretical false positive detection rate to a suitably low level. Then only two of the tracks could have been detected, and in practice there would have been many false positives flagged.

| FP | FP% | FN | FN% | DET | TOT |
|---|---|---|---|---|---|
| 60 | 0.7 | 14 | 0.0033 | 8539 | 429238 |

Table 2: The number of false positive objects (FP) and false negatives (FN) for satellite track/scratch detection. False positive percentage is expressed as percentage of total object detections (DET); false negative percentage is expressed as percentage of [total objects (TOT) - total detections (DET)].

A general summary of the results can be found in table 2. All the major satellite tracks were found and the ends of the tracks were generally accurately delineated. All the small scratches were properly identified, although one involved a significant bend. Figure 4 illustrates this. Some of the objects along the bend were improperly classified as real objects. A small number of false positive detections of short lines was made.

An example of a very small scratch that was found can be seen in Figure 5. This figure also shows a num-ber of points which were flagged as spurious by the renewal string approach. This is located on UKR001 around $RA = 2:34:52$, $DEC = -86:32:16$. The astronomers marked this as a set of false positive detections, as it appeared to look like stars and galaxies which just happened to be aligned. However it is clear from the image (figure 5c) that these points are aligned along a very faint track. The renewal string approach is picking up objects which would not be recognised by the human eye looking at the data alone.

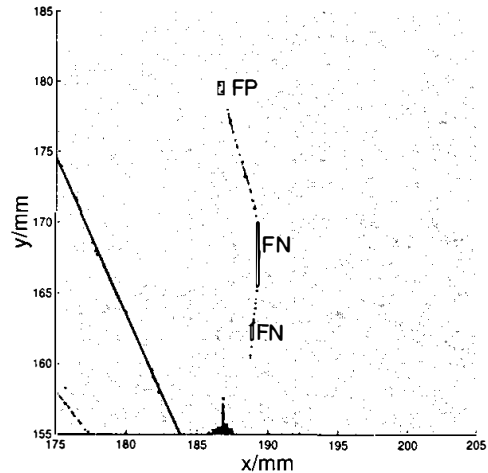

Figure 4: False positives (FP) and false negatives (FN) for detections along a very faint highly curved scratch on plate UKR001. This is the only significant source of false negatives for scratch/track detection that the astronomers found on this plate (the others were isolated points at the end of a scratch).

## 6  DISCUSSION

Renewal Strings have certainly aided the process of detection of spurious objects in astronomical data: given very large amounts of data only a small number of detections were made, most of which were correct. The form of the model allows the use of the hidden Markov models and renewal processes, resulting in a model that is efficient even for huge datasets. It has been run all the plates of the SSS data (over 3000 in total), providing a valuable resource to astronomers.

Renewal strings are a practical, probabilistic approach to a large problem requiring high accuracy. Renewal strings go beyond a local Hough transform method to a general approach for detecting line segments within large amounts of other data. Slightly curved lines are also detectable as a set of locally linear parts.

The renewal string approach shows clear benefits over Hough approaches, and has proven a highly effective method for detection of spurious data in the Supercosmos Sky Surveys. The result of the method will reduce the problem of spurious data to insignificant levels. Furthermore the technique is general and can be adapted for use in future sky surveys.

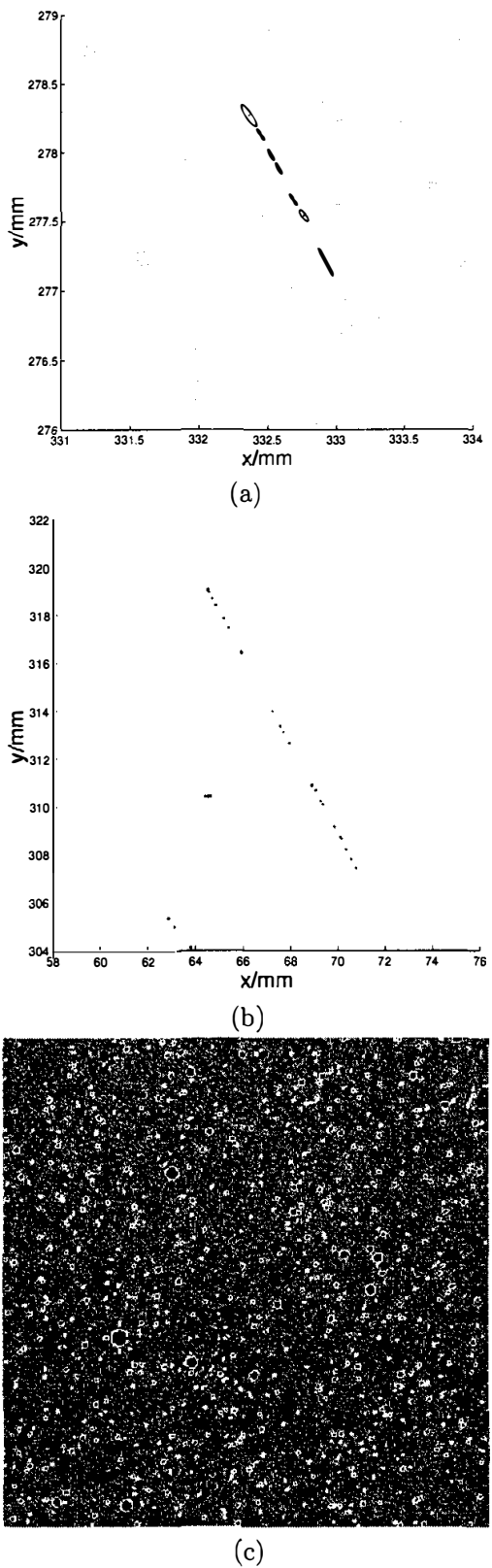

Figure 5: (a) A very short scratch which was detected. (b) A set of detections on UKR001 which were marked up as false positives by the astronomers. (c) A look at the image shows the points are in fact part of a faint track.

### Acknowledgements

This work is part of a project funded by the University of Edinburgh. The authors also thank IBM for the provision of a p-series machine, Blue Dwarf, to the School of Informatics, Edinburgh.